\documentclass{article}

\PassOptionsToPackage{numbers, compress}{natbib}

\usepackage[preprint]{neurips_2024}

\usepackage[utf8]{inputenc} 
\usepackage[T1]{fontenc}    
\usepackage{hyperref}       
\usepackage{url}            
\usepackage{booktabs}       
\usepackage{amsfonts}       
\usepackage{nicefrac}       
\usepackage{microtype}      
\usepackage{xcolor}         

\title{Formatting Instructions For NeurIPS 2024}

\usepackage{amssymb}
\usepackage{amsmath}
\usepackage{mathabx}

\usepackage{microtype}

\usepackage{graphicx}
\usepackage{xcolor}
\usepackage{xspace}

\usepackage{booktabs}
\usepackage{multirow}

\usepackage[labelformat=simple]{subcaption}
\usepackage{caption}
\input{Definitions}

\newcommand{\mwoz}{MultiWoZ 2.4\xspace}
\newcommand{\acibench}{ACI-Bench\xspace}
\newcommand{\radqa}{RadQA\xspace}

\newcommand{\llama}{LLaMA2\xspace}

\newcommand{\pie}{\textsc{PiE}\xspace}
\newcommand{\pid}{\textsc{PiD}\xspace}
\newcommand{\sys}{\textsc{PiD}\xspace}

\newcommand{\baseM}{$_\text{base}$\xspace}
\newcommand{\largeM}{$_\text{large}$\xspace}
\newcommand{\flops}{FLOPs\xspace}
\newcommand{\codex}{Codex$_{\text{davinci}}$\xspace}

\newcommand{\ensuretext}[1]{#1}

\newcommand{\arkcomment}[3]{\ensuretext{\textcolor{#3}{[#1 #2]}}}
\renewcommand{\arkcomment}[3]{}  

\definecolor{green}{RGB}{0,128,0}
\definecolor{lightgray}{RGB}{211, 211, 211}
\definecolor{pink}{RGB}{255,0,255}
\definecolor{red}{RGB}{219,68,55}
\definecolor{yellow}{RGB}{244,180,0}
\definecolor{blue}{RGB}{66,133,244}
\definecolor{lightyellow}{RGB}{255,255,0}

\title{Efficient Encoder-Decoder Transformer Decoding \\ for Decomposable Tasks}

\DeclareSymbolFont{extraup}{U}{zavm}{m}{n}
\DeclareMathSymbol{\varheart}{\mathalpha}{extraup}{86}
\DeclareMathSymbol{\vardiamond}{\mathalpha}{extraup}{87}

\newcommand\uw{$^\spadesuit$}
\newcommand\msr{$^{\heartsuit}$}

\newcommand\ai{$^{\diamondsuit}$}
\newcommand\aspace{\hspace{.75em}}

\author{
Bo-Ru Lu\uw\aspace
Nikita Haduong\uw\aspace
Chien-Yu Lin\uw\aspace
\textbf{Hao Cheng}\msr\aspace \\
\textbf{Noah A. Smith}\uw\ai\aspace
\textbf{Mari Ostendorf}\uw\aspace\\\\
\uw University of Washington \aspace
\msr Microsoft Research \aspace
\ai Allen Institute for AI\\\\
{\tt roylu@washington.edu}
}

\begin{document}
\maketitle
\begin{abstract}

Transformer-based NLP models are powerful but have high computational costs that limit deployment. Finetuned encoder-decoder models are popular in specialized domains and can outperform larger more generalized decoder-only models, such as GPT-4.  We introduce a new configuration for encoder-decoder models that improves efficiency on structured output and decomposable tasks where multiple outputs are required for a single shared input. Our method, \textbf{\textit{prompt-in-decoder}} (\sys), encodes the input once and decodes the output in parallel, boosting both training and inference efficiency by avoiding duplicate input encoding and increasing the operational intensity (ratio of numbers of arithmetic operation to memory access) of decoding process by sharing the input key-value cache. We achieve computation reduction that roughly scales with the number of subtasks, gaining up to 4.6x speed-up over state-of-the-art models for dialogue state tracking, summarization, and question-answering tasks, with comparable or better performance.
\end{abstract}

\section{Introduction}

The transformer architecture \cite{vaswani2017attention} is the backbone of many successful NLP models, but they have
high latency and computational costs.
To reduce costs, researchers have investigated multiple approaches, including: 
i)~model compression (e.g., distillation \cite{hinton2015distilling,gou2021knowledge,Udagawa2023ACA}, quantization \cite{zadeh2020gobo,dettmers2022gptint,yao2022zeroquant,dettmers2023qlora,zhao2023atom}, and mixture of experts \cite{kudugunta-etal-2021-beyond-distillation}); 
ii)~decoding strategy (e.g., speculative decoding \cite{leviathan2023fast, chen2023accelerating} and parallel decoding \cite{santilli-etal-2023-accelerating, ning2023skeleton}); 
iii)~algorithm-level attention optimizations (e.g., sparse attention \cite{10.1162/tacl_a_00353,Liu2022DynamicSA}, fewer number of key-value heads \cite{Shazeer2019FastTD,ainslie-etal-2023-gqa} and Hydragen \cite{juravsky2024hydragen}); 
iv)~kernel-level attention optimizations (e.g., FlashAttention \cite{dao2022flashattention} and FlashInfer \cite{flashinfer}); 
and v)~memory management (e.g. paged attention\cite{kwon2023efficient} and radix attention \cite{zheng2023efficiently}).
Our work falls into the category of algorithm-level attention optimizations via key-value prefix sharing.
We aim to upcycle existing encoder-decoder models with the goal of reducing redundant computations and increasing hardware utilization via key-value prefix sharing.

\begin{figure}
  \begin{subfigure}[t]{\linewidth}
    \centering
    \includegraphics[width=\linewidth]{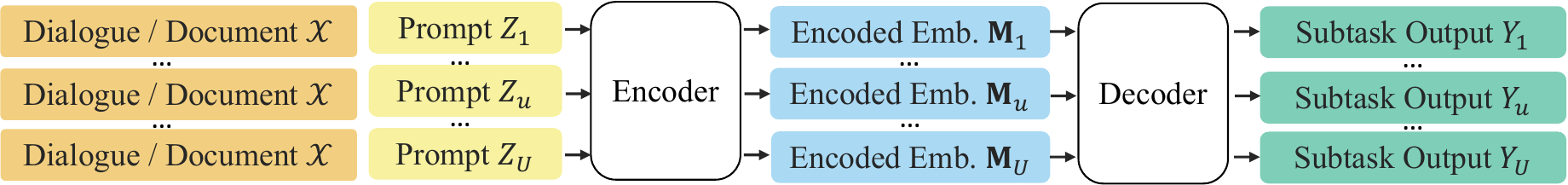}
    \caption{Prompt-in-encoder (\pie).}
    \label{fig:pie}
  \end{subfigure}%
  \par\bigskip
  \begin{subfigure}[t]{\linewidth}
    \centering
    \includegraphics[width=\linewidth]{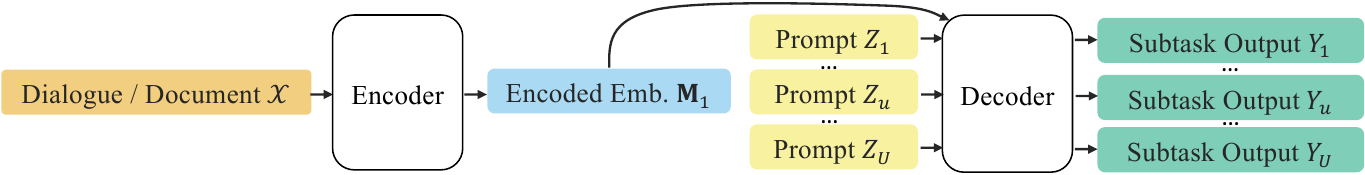}
    \caption{(Ours) Prompt-in-decoder (\sys).}
    \label{fig:sys}
  \end{subfigure}
  \caption{Given a task where a single input document $\Xcal$ is used to generate multiple outputs $Y_u$ associated with different prompts $Z_u$, \pie creates unique encodings $\Mb_u$ for every prompt $Z_u$. In contrast, \sys uses a single shared $\Mb$ for each prompt. thus requiring less memory access and resulting in  higher computational efficiency.}
  \label{fig:teaser}
\end{figure}

Recent research \cite{wei2022emergent} indicates that scaling up decoder-only transformer models facilitates effective in-context learning, which is particularly beneficial when labeled data is scarce, making the models more generalized to unseen domains. However, these larger decoder-only models, despite their generalizability to unseen domains, incur higher computational costs and may offer inferior performance compared to smaller, specialized encoder-decoder transformer models \cite{pmlr-v209-eric23a}. 
Motivated by this, we propose a new configuration for encoder-decoder models to further improve training and inference efficiency in \emph{decomposable} tasks that involve multiple prompts over the same document (or dialogue).  
These tasks include scenarios where multiple users are querying the same document with different requests (e.g., question answering, \cite{cascade-inference,juravsky2024hydragen}), as well as scenarios where it is useful to decompose a complex task into simpler subtasks (e.g., abstractive summarization of long/multiple documents/dialogues \cite{gidiotis2020divide,meng-etal-2021-bringing,zhang-etal-2022-summn,yim2023aci} and dialogue information extraction \cite{lu2023dialgen}). 

In the multi-user question-answering scenario, decoupling questions and the corresponding document allows reusing the shared document embeddings for questions from different users. 
The shared embeddings can significantly reduce duplicate computation of the shared prefixes and increase operational intensity during training and inference. 
For summarization and information extraction, decomposing a long target output into multiple shorter sequences mitigates attention degeneration issues \cite{fu2023decoder,zhou-etal-2023-building}, leading to a boost in accuracy and efficiency. 
However, existing methods put the prompts in the encoder, resulting in a high computation cost because contextualizing each shared document and its prompts results in the duplicate encoding of the same shared prefixes.
Instead, we propose \textbf{prompt-in-decoder (\sys)}: an encode-once, decode-in-parallel strategy that avoids duplicate encoding costs by sharing inputs and increases decoding efficiency by reducing memory access.

We demonstrate the effectiveness of our decoding strategy through experiments on a range of tasks with short and long inputs: dialogue state tracking, abstractive medical dialogue summarization, and extractive medical question answering. Our models achieve comparable or higher performance (98-101\%) than the current state of the art. At the same time, we observe a 2-10x computation reduction, depending on the number of subtasks, and up to 4.6x speed-up for shorter subtask outputs.

In summary, the main contribution of this work is a new, \textbf{more efficient decoding configuration for encoder-decoder models on decomposable tasks}, validated on multiple language tasks, including dialogue state tracking, medical summarization and question answering. 
\section{Encoder-Decoder Framework}
\label{sec:background}

The encoder-decoder is a general framework that has been used to address a wide variety of problems in NLP \citep{lewis-etal-2020-bart,JMLR:v21:20-074}.  Given an input word sequence, a desired result is obtained by first encoding the input and then iteratively generating an output word sequence.
In general-purpose models, the input $\Xcal$ is optionally combined with a prompt
$\Zcal$ that specifies the task: $\Ycal = \mathrm{decoder}(\mathrm{encoder}(\Xcal , \Zcal)).$
The input $\Xcal$ is any form of text, e.g., a sentence, article, or transcript of a conversation, and $\Zcal$ can be an instruction or a question.
The output $\Ycal$ could be information extracted from an article, a summary of a conversation, or a response to a question.
State-of-the-art encoder-decoder systems are built on transformers. This section overviews the general framework to introduce notation and set up the multi-subtask inference problem that we address.

\subsection{Multi-Prompt Decoding}
\label{sec:PiE}
In this paper, we tackle tasks that can be framed in terms of multiple prompts over the same input $\Xcal$. 
Specifically, the output is a list of subtasks (or answers) $\Ycal$, where each subtask/answer is $Y_u$, e.g., $\Ycal = (Y_1, \dots, Y_u, \dots, Y_U)$, and $U$ is the total number of subtasks/answers. Each subtask $Y_u$ is associated with a specific prompt $Z_u$, so
$\Zcal = (Z_1, \dots, Z_u, \dots, Z_U)$. 
The scenario involves running inference multiple times to generate $Y_u$, 
$Y_u = \mathrm{decoder}(\mathrm{encoder}(\Xcal, Z_u))$, then combining all outputs $Y_u$ to form the final $\Ycal$.
We refer to this as the \textbf{prompt-in-encoder (\pie)} approach. 
\autoref{fig:pie} illustrates how \pie tackles a single instance ($\Xcal$, $\Ycal$ and $\Zcal$).
\pie involves $U$ encodings of $\Xcal$, one for each prompt $Z_u$.

\subsection{Encode Once and Decode in Parallel}

To avoid the redundant encoding of $\Xcal$ in \pie and improve inference efficiency when decoding $Y_u$, we propose placing prompt $Z_u$ in the decoder, allowing us to encode $\Xcal$ once and decode $Y_u$ in parallel. We refer to this method as \textbf{prompt-in-decoder (\sys)}.
By moving $Z_u$ from the encoder (in \pie) to the decoder, the encoder only encodes $\Xcal$ once, generating a single sequence of embeddings that is reused throughout the decoding process for each prompt $Z_u$.
As shown in \autoref{fig:sys}, $\Xcal$ is only encoded once, and the embeddings $\Mb$ are reused for $U$ prompts during decoding to generate all subtask outputs $(Y_1, \dots, Y_u, \dots, Y_U)$. 
Formally, the equation can be represented as $Y_u = \mathrm{decoder}(\mathrm{encoder}(\Xcal), Z_u).$

\section{Performance Analysis}
\label{sec:performance_analysis}

In this section, we show how the \sys model improves inference efficiency over the \pie model by quantifying memory access and the number of arithmetic operations. 

\subsection{Operational Intensity}
The operation intensity is the number of operations per byte of memory access, expressed as:
\newcommand\myeq{\mkern1.5mu{=}\mkern1.5mu}
\[
   \text{Operational Intensity} = \frac{\text{FLOP/s}}{\text{byte/s}} = \frac{\text{\# arithmetic operations}}{\text{memory access}}\text{,}
\]
which provides a measure for the hardware efficiency \cite{10.1145/1498765.149878} or algorithm efficiency.
To carry out calculations, accelerators must access and move data between global memory and registers, which can be a bottleneck because modern hardware accelerators such as GPUs/TPUs often have significantly greater capacity for computations compared to memory bandwidth. 
For example, an NVIDIA A100 GPU \cite{choquette2021nvidia} has an operation capacity of 312 Tera FLOP/s versus a memory bandwidth of 2 Gigabyte/s. 
If the operational intensity is too low, the accelerator idles, waiting for data to move to registers instead of running computations. This often occurs in models where memory access is more intensive than arithmetic operations, i.e., incremental decoding in transformers \cite{Shazeer2019FastTD}. 
By decreasing memory access, the operational intensity is increased, i.e., efficiency improves.  

\begin{figure}[t!]
  \begin{subfigure}[t]{0.47\textwidth}
    \centering
    \includegraphics[width=\linewidth]{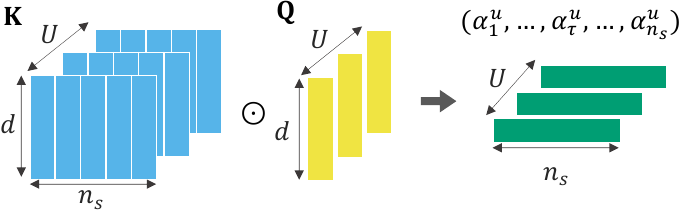}
    \caption{Cross-attention of \pie.}
    \label{fig:pie_cross-attn}
  \end{subfigure}
  \hspace{3mm}
  \begin{subfigure}[t]{0.47\textwidth}
    \centering
    \includegraphics[width=\linewidth]{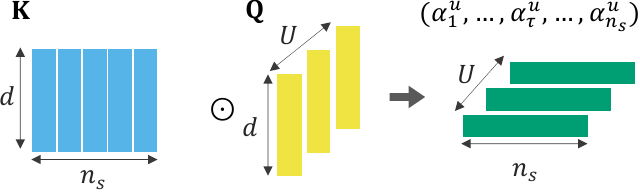}
    \caption{Cross-attention of \sys.}
    \label{fig:sys_cross-attn}
  \end{subfigure}
  \caption{An illustration of cross-attention dot product operations ($\Qb \Kb^\top$ in \autoref{eq:attn}) for \pie and \sys for a single inference step. $U$, $d$, $n_s$ are the number of prompts, hidden layer dimension, and input length, respectively. $\odot$ is the dot product operation, and the resulting scalars of $\Qb \Kb^\top$ are $\alpha^u_\tau$, where $\tau = \{1,\ldots, n_s\}$, at the decoding step $\tau$ w.r.t.~the prompt $Z_u$. 
  }
  \label{fig:cross-attn}
\end{figure}

\subsection{Multi-head Attention in Transformer}
\label{sec:defns}
Transformers~\cite{vaswani2017attention} have two types of multi-head attention: self-attention and cross-attention. Usually, the attention key and value tensors have the dimension $d/h$, where $d$ is the dimension of input and output vectors and $h$ is the number of attention heads. For simplicity, we consider the operations of all heads together such that the dimension of $\Qb$, $\Kb$, $\Vb$ (query/key/value) in the following section is denoted as $d$. 

During attention, the query/key/value vectors can be obtained by projecting the corresponding input vectors $\Nb \in \R^{n \times d} $ or $\Mb \in \R^{m \times d}$, where $n$ and $m$ can be either $n_s$ (input source length) or $n_t$ (output target length). More formally, the equations are $\Qb = \Nb \cdot W^Q \in \R^{n \times d}$, $\Kb = \Mb \cdot W^K \in \R^{m \times d}$ and $\Vb = \Mb \cdot W^V \in \R^{m \times d}$ where the projection matrices are $W^Q$, $W^K$, $W^V \in \R^{d \times d}$.

In the self-attention, $\Nb$ is equivalent to $\Mb$; hence, $m = n = n_s$ in the encoder or $m = n = n_t$ in the decoder. On the other hand, in the case of the cross-attention, the variable $\Nb \in \R^{n_t \times d}$ originates from the decoder, while $\Mb \in \R^{n_s \times d}$ is sourced from the output of the encoder.

The simplified equation of the attention mechanism is represented as follows,
\begin{equation}
\label{eq:attn}
\Ob = \mathrm{softmax}\left(\frac{\Qb \Kb^\top}{\sqrt{d/h}}\right) \Vb \cdot W^O\text{,}
\end{equation}
where $\Ob \in \R^{m \times d}$ is the final atention output.

\label{sec:performance}

\subsection{Performance Analysis for \pie and \sys}

\autoref{fig:cross-attn} shows the dot product operation in the cross-attention for the two models. 
In the \pie model, the input is contextualized with each prompt, so the encoder's output tensor differs for each prompt.
Thus, at each decoding step $\tau$, $\Kb$ is read $U$ times to generate $U$ different sets of attention weights $\alpha^u_{\tau}$ to decode $Y_u$.
In contrast, in the \sys model,  $\Kb$ is shared across all prompts since the input is encoded independently of the prompts. Thus, at each decoding step $\tau$, the dot product operation in the cross-attention shares and broadcasts $\Kb$ and computes the dot product of $\Kb$ and $\Qb$, resulting in lower memory access but the same number of arithmetic operations compared to \pie.

We approximate the memory access and operations based on the dominant terms in the self- and cross-attention and ignore the constant terms. For a single input, the memory access of $\Mb$ and $\Nb$ is $n_s d$ and $n_t d$. The memory access of the matrices $W^Q$, $W^K$, $W^V$, $W^O$ is $d^2$. The number of operations in both attention mechanisms is dominated by the matrix projections which are used to obtain $\Qb$, $\Kb$, $\Vb$, and $\Ob$; thus, the number of operations is approximated as $n_s d^2$ or $n_t d^2$.

\begin{table*}[!htp]
    \centering
    \footnotesize
    \caption{Inference computation comparison between \pie and \sys, where $U$, $b$, $n_s$, $n_t$, $d$ are the number of prompts, batch size, input source length, output target length, and hidden size, respectively.}
    \resizebox{\linewidth}{!}{
        \begin{tabular}{ccccccc}
        \toprule
             & \multicolumn{2}{c}{Encoder's Self-attention}
             & \multicolumn{2}{c}{Decoder's Self-attention}
             & \multicolumn{2}{c}{Decoder's Cross-attention} \\
              \cmidrule(lr){2-3}
              \cmidrule(lr){4-5}
              \cmidrule(lr){6-7}
             Model
             & Memory & Operations 
             & Memory & Operations 
             & Memory & Operations \\\midrule 
             \pie-T5  & $Ub n_s d + d^2$ & $Ub n_s d^2$ 
                      & $Ub n^2_t d + n_t d^2$ & $Ub n_t d^2$ 
                      & $Ub n_s n_t d + Ub n_t d + n_t d^2 $ & $Ub n_t d^2$ \\\midrule 
             \sys-T5  & \colorbox{yellow!50}{$b n_s d$}$+d^2$  & \colorbox{yellow!50}{$b n_s d^2$}  
                      & $Ub n^2_t d + n_t d^2$ & $Ub n_t d^2$ 
                      & \colorbox{yellow!50}{$b n_s n_t d$}$+ Ub n_t d + n_t d^2 $  & $Ub n_t d^2$ \\
        \bottomrule
        \end{tabular}
    }
    \label{tab:performance_analysis}
\end{table*}
The comparisons of memory access and operation counts on different inference components for our \pie and \sys implementations are presented in~\autoref{tab:performance_analysis}.
The table demonstrates that \sys enhances efficiency by encoding only once and by boosting decoding efficiency through reduced memory access, achieved by sharing the input key-value cache.
The count analysis is explained in further detail below.
In the analysis, we assume that a batch of $b$ inputs with the same set of $U$ subtasks are processed together.
The encoder input length and the decoder output length differ depending on whether the prompt is in the decoder. For simplicity, the analysis also assumes that the prompt terms are negligible; Appendix~\ref{app:comp-anal} provides the detailed justification. 

\paragraph{Encoder's self-attention.}
In \pie, to run inference on a single instance, the model encodes $U$ prompts ($Z_u$) with input ($\Xcal$). Considering a batch with $b$ instances, \pie takes $U b n_s d + d^2$ for memory access and $U b n_s d^2$ for the number of operations. In contrast, \sys only encodes the input once, so the memory access and the number of operations remain $b n_s d + d^2$ and $b n_s d^2$. Thus, the encoders of both models have similar operational intensity, but \pie requires more memory access and more arithmetic computations. 

\paragraph{Decoder's self-attention.}
In both \pie and \sys, the decoder computes the self-attention for $U$ prompts. For each decoding step, the memory access and the number of operations are $U b n_t d$ and $U b d^2$. Thus, for $n_t$ steps, the resulting memory access and the number of operations are $U b n^2_t d$ and $U b n_t d^2$. In this case, \pie and \sys have roughly the same operational intensity.

\paragraph{Decoder's cross-attention.}
Cross-attention dominates the inference cost.
For each decoding step, we load the encoded input embeddings $\Mb \in \R^{m \times d}$, where $m = n_s$, from the encoder for each $(\Xcal ,Z_u)$ input for \pie and $\Xcal$ for \sys. For \pie, 
the resulting $b$ $(\Mb_1, \dots, \Mb_U)$ are fed into the decoder's cross-attention for $b$ instances with $U$ prompts. On the other hand, \sys shares all $\Mb$ for all $U$ prompts of each instance, resulting in feeding $b$ $\Mb$ to the decoder. Therefore, for each step, the memory access for loading encoded $\Mb$ is $U b n_s d $ and $b n_s d$ for \pie and \sys, respectively. The memory access of loading $\Qb$ and $\Ob$ is $U b d$. Loading projection matrices takes $d^2$. We multiply all memory access cost by a factor of $n_t$ steps.
\section{Datasets \& Metrics}
\label{sec:datasets}

\subsection{Datasets \& Task Performance Metrics}
\paragraph{Dialogue State Tracking (DST).} Multi-domain Wizard-of-Oz dataset (MultiWoZ)\cite{budzianowski-etal-2018-multiwoz} is a task-oriented dialogue dataset.
We selected the most recent version of \mwoz \cite{ye-etal-2022-multiwoz} due to its refined validation and test set annotations.
For comparison to other work, joint goal accuracy (JGA) is adopted as the evaluation metric. 
The input $\Xcal$ is the dialogue history, $\Ycal$ is the dialogue state, the subtask prompts $Z_u$ are the domain-slot name, and the associated outputs $Y_u$ are slot values.

\paragraph{Summarization.}
We use \acibench\cite{yim2023aci}, a dataset containing clinical notes associated with conversations between doctors and patients. The clinical notes have structured output with distinct sections. 
We use ROUGE-L score \cite{lin2004rouge}, denoted as R-L, to evaluate models.
The input $\Xcal$ is the doctor-patient dialogue, $\Ycal$ is the full clinical note, the subtask prompts $Z_u$ are section indicators, and the associated outputs $Y_u$ are section notes.

\paragraph{Question Answering.} 
\radqa \cite{soni-etal-2022-radqa} is an extractive question-answering dataset on radiology reports with 3k questions posed by experts. A single report can have multiple questions. 
We use exact match (EM) as our evaluation metric.
The input $\Xcal$ is the radiology report, the subtask prompts $Z_u$ are specific questions, and the associated outputs $Y_u$ are extracted responses.

\subsection{Efficiency Metrics}

\paragraph{Floating point operations (\flops)}
refer to the number of arithmetic operations required for model inference, i.e., the computational complexity.\footnote{We use \texttt{calflops}~\citep{calflops} to compute \flops.} 
Note that FLOP reduction may not
correlate with wall-clock speed, as it tends to ignore overheads from memory access (IO) \cite{dao2022flashattention}.

\paragraph{Latency.} To account for extra IO costs, e.g., GPU memory bandwidth, we also report latency, which measures the wall clock time for a single instance inference in a single-user scenario (e.g., an edge computing device). Specifically, we report the average time for a single instance inference, where instances are processed sequentially.

\paragraph{Latency w/ Batching.} For cloud-serving applications, multiple instances are computed in a batch fashion to fully utilize the computing device. We thus report the average time for a single instance, where a batch of instances is computed simultaneously.

To assess the \flops and latency, we randomly chose 512 samples from \mwoz test set (due to the large test set) and used the full test sets for \acibench and \radqa. 
We report the average latency and the average latency w/ batching at the optimal batch size.
\section{Experiments \& Results}

\subsection{Compared Systems}
\paragraph{T5}\cite{JMLR:v21:20-074} is adopted as the encoder-decoder model in all experiments. 
We use T5-base and T5-large models from HuggingFace for \mwoz and \acibench.  For \radqa, we follow the previous work \cite{pmlr-v209-eric23a} and use a pretrained clinical T5.
We denote base and large models in following tables as T5\baseM and T5\largeM (220M and 770M parameters respectively). 
We use T5 as the backbone to compare two different subtasking strategies: prompt-in-encoder (\pie) and our proposed prompt-in-decoder (\sys); thus the model size keeps the same as the standard T5.

\paragraph{\llama} \citet{touvron2023llama} is a popular open decoder-only language model. Due to computation limitations, we adopt low-rank adaptation (LoRA) \cite{hu2022lora} to efficiently finetune \llama 7B, resulting in approximately 80M trainable parameters with a rank of 64. 

\paragraph{Current state-of-the-art models.} 
We report the current best published results for in-context learning and full finetuning for every dataset. The source papers for the results of each dataset are documented in the caption of \autoref{tab:sota}.

\subsection{Training Procedure}
\label{subsec:training_procedure}
The standard fine-tuned T5 and \llama data is represented as $(\Xcal, \Ycal)$, i.e.\ no prompts are used. For \pie-T5, the data is $(\Xcal, Y_u, Z_u)$, since $\Xcal$ is separately contextualized with each subtask prompt $Z_u$ and a subtask output $Y_u$, effectively increasing the dataset size by a factor of $U$ and substantially extending the time required for training. In contrast, \sys-T5 has the flexibility to be trained using either data representation, as it uses shared inputs. We choose $(\Xcal, \Ycal, \Zcal)$ for \sys-T5 because it is more efficient to put all output $\Ycal$ from the same shared input $\Xcal$ in the same batch to save encoding processing time during gradient update. 
The model selection criterion is the highest score on the validation set. Hyperparameters can be found in \autoref{app:training_details}.

\subsection{Results}

\paragraph{Task performance.}
Table \ref{tab:task_performance} presents the task performance results for three different full finetuning scenarios (T5, \pie-T5, and \sys-T5), as well as a LoRA finetuned LLaMA2 language model.
The results of \mwoz and \acibench indicate that subtasking and multi-prompt decoding help, since both \pie-T5 and \sys-T5 outperform the standard T5 and \llama models.
Although larger model size often comes with better performance in moving from T5\baseM to T5\largeM, there are mixed results for \llama on these domains. One possible reason could be that the number of trainable parameters is fewer than that of full finetuning. 
More importantly, similar or greater gains in performance are obtained with smaller, more efficient models that leverage task structure.
\begin{table}[h]
    \caption{Task performance comparison between the baselines and \sys-T5 over the three test sets.}
    \label{tab:task_performance}
    \centering
    \small
    \begin{tabular}{cccc}
        \toprule
             & \mwoz & \acibench & \radqa \\
             Model 
             & JGA   $\uparrow$ & ROUGE-L $\uparrow$ & EM $\uparrow$ \\
        \midrule
        T5\baseM                 & 71.5$_{\pm0.4}$ & 47.9$_{\pm0.1}$ & --   \\
        \pie-T5\baseM            & 76.3$_{\pm0.7}$ & 53.8$_{\pm0.4}$ & 53.7$_{\pm0.5}$ \\
        \sys-T5\baseM            & 75.5$_{\pm0.4}$ & 54.0$_{\pm0.5}$ & 52.4$_{\pm0.6}$ \\\midrule
        T5\largeM                & 72.5$_{\pm0.7}$ & 52.5$_{\pm0.3}$ & --  \\
                                        
        \pie-T5\largeM           & 77.5$_{\pm0.8}$ & 54.7$_{\pm0.6}$ & 55.4$_{\pm0.3}$ \\
        \sys-T5\largeM           & 76.5$_{\pm0.4}$ & 55.2$_{\pm0.7}$ & 54.6$_{\pm0.4}$ \\
        \llama$_\text{7B}$       & 66.1$_{\pm0.1}$ & 53.8$_{\pm0.6}$ & 54.4$_{\pm1.0}$ \\
        \bottomrule
        \end{tabular}
\end{table}
\definecolor{orange}{HTML}{E66100}
\definecolor{blue}{HTML}{0C7BDC}
\definecolor{yellow}{HTML}{FFC20A}
\newcommand{\FS}[1]{\cellcolor{blue!25} \text{{#1}}}
\newcommand{\SL}[1]{\cellcolor{yellow!25} \text{{#1}}}
\newcommand{\ST}[1]{\cellcolor{orange!25} \text{{#1}}}

\begin{table*}[!htp]
    \caption{ 
    Task performance and inference efficiency comparison between \pie-T5, \sys-T5 and \llama over three test sets.
    Sp and Sp$^{\text{batch}}$ represent the relative speed-up in single-instance and batching scenarios computed on NVIDIA A100.
    We present the relative ratios of \flops, Sp and Sp$^{\text{batch}}$ compared to \pie-T5\baseM or \pie-T5\largeM, across other models. 
    Overall, \sys-T5 achieves best computation efficiency across all tasks and achieves comparable task performance on \mwoz and \radqa and better task performance on \acibench.}
    \centering
    \resizebox{\linewidth}{!}{
        \begin{tabular}{ccccccccccccc}
        \toprule
             & \multicolumn{4}{c}{\mwoz}
             & \multicolumn{4}{c}{\acibench}
             & \multicolumn{4}{c}{\radqa} \\
              \cmidrule(lr){2-5}
              \cmidrule(lr){6-9}
              \cmidrule(lr){10-13}
             Model 
             & JGA   $\uparrow$ & \flops $\downarrow$   & Sp$\uparrow$ & Sp$^{\text{batch}}$ $\uparrow$ 
             & R-L   $\uparrow$ & \flops $\downarrow$   & Sp$\uparrow$ & Sp$^{\text{batch}}$ $\uparrow$ 
             & EM    $\uparrow$ & \flops $\downarrow$   & Sp$\uparrow$ & Sp$^{\text{batch}}$ $\uparrow$\\
        \midrule
        \pie-T5\baseM         & 76.3 & 1.0x & 1.0x & 1.0x 
                              & 53.8 & 1.0x & 1.0x & 1.0x 
                              & 53.7 & 1.0x & 1.0x & 1.0x \\
        \sys-T5\baseM         & 75.5 & \bf0.1x & \bf1.9x & \bf4.6x
                              & 54.0 & \bf0.4x & \bf1.1x & \bf1.1x
                              & 52.4 & \bf0.6x & \bf1.3x & \bf1.3x\\\midrule
                              
        \pie-T5\largeM        & 77.5 & 1.0x & 1.0x & 1.0x
                              & 54.7 & 1.0x & 1.0x & 1.0x
                              & 55.4 & 1.0x & 1.0x & 1.0x \\
        \sys-T5\largeM        & 76.5 & \bf0.1x & \bf2.8x & \bf4.2x
                              & 55.2 & \bf0.4x & 1.0x & \bf1.5x
                              & 54.6 & \bf0.5x & \bf2.8x & \bf1.3x \\
                              
        \llama$_\text{7B}$  & 66.1 & 0.7x & 0.2x & 0.5x
                              & 53.8 & 4.8x & 0.3x & 0.3x
                              & 54.4 & 26.8x& 0.2x & 0.1x  \\
        \bottomrule
        \end{tabular}
    }
    \label{tab:computation_cost}
\end{table*}
\paragraph{Computation efficiency.}
Since \pie-T5 and \pid-T5 achieve better results than standard T5, we compare \pie-T5 and \pid-T5 in \autoref{tab:computation_cost}.
Regarding computational efficiency, measured in \flops, \sys-T5 significantly outperforms \pie-T5 in reducing the number of arithmetic operations because it processes each input only once. \sys-T5 achieves superior speed-up in both single-instance and batching scenarios across three datasets and two model sizes, while obtaining similar performance (98-101\%) to \pie-T5.
Additionally, \sys-T5 offers greater reductions in computational costs (2-10x) and further accelerates efficiency when dealing with a larger number of subtasks—for example, managing 30 slots in \mwoz versus 4 sections in \acibench and 2-6 questions in \radqa.

\begin{table*}[htbp]
    \caption{We choose previous state-of-the-art (SOTA) generative models from the literature with the most comparable model sizes. In-context learning (ICL) SOTA results for \mwoz, \acibench, and \radqa are reported in the studies \cite{hu-etal-2022-context, yim2023aci, pmlr-v209-eric23a}. For full finetuning (FT) SOTA, the results are reported in the studies \cite{zhao2022description, yim2023aci, pmlr-v209-eric23a}. \citet{yim2023aci} use four Bart\largeM models (one for each section), resulting in quadruple the size of a single Bart\largeM (406M). All numbers are reported on the test sets.}
    \centering
    \small
    \resizebox{\linewidth}{!}{
        \begin{tabular}{cccccccccccc}
        \toprule
             & \multicolumn{3}{c}{\mwoz}
             & \multicolumn{3}{c}{\acibench}
             & \multicolumn{3}{c}{\radqa} \\
              \cmidrule(lr){2-4}
              \cmidrule(lr){5-7}
              \cmidrule(lr){8-10}
             & Model & Size$\downarrow$ & JGA   $\uparrow$ 
             & Model & Size$\downarrow$ & R-L   $\uparrow$ 
             & Model & Size$\downarrow$ & EM    $\uparrow$  \\
        \midrule
        ICL SOTA
        & \codex              & 175B           & 62.4 
        & GPT4-32k            & 1760B          & 54.3
        & GPT3                & 175B           & 36.2 \\
        FT SOTA  
        & T5$_\text{XXL}$     &  11B           & 75.9
        & Bart\largeM         & 4$\times$406M  & 48.6 
        & T5\largeM           & \bf770M        & \bf55.0 \\
        FT (Ours)      
        & \sys-T5\largeM           & \bf770M        & \bf76.5
        & \sys-T5\largeM           & \bf770M        & \bf55.2
        & \sys-T5\largeM           & \bf770M        & 54.6   \\
        \bottomrule
        \end{tabular}
    }
    \label{tab:sota}
\end{table*}
\paragraph{Comparison between \sys-T5\largeM and state-of-the-art models.}
\autoref{tab:sota} illustrates the comparison between our method, \sys-T5\largeM, and existing state-of-the-art  approaches. 
Our \sys-T5\largeM outperforms in-context learning methods on all three datasets with much smaller models.
Compared to full fine-tuning, our model outperforms on \mwoz and \acibench, but underperforms on \radqa. This discrepancy could be attributed to the fact that the prompts in \radqa are less structured compared to the fixed types of prompts in tasks with structured outputs.

We omit the inference cost of state-of-the-art models since access to the language models for in-context learning is restricted to API calls, and some finetuned models' checkpoints are unavailable. Nevertheless, we can assume that the inference cost exceeds that of the proposed \sys-T5 due to the larger model sizes or inference techniques employed. The T5$_{\text{XXL}}$ inference in \mwoz follows the standard T5, while the inference approaches for Bart\largeM and T5\largeM in \acibench and \radqa are consistent with those in \pie-T5.

\begin{table}[!htp]
    \caption{Comparison between in-context learning and full finetuning on \mwoz test set. The JGA scores are reported at 1\%, 5\%, 10\% and 100\% of training data.
    }
    \label{tab:icl_comparison}
    \centering
    \small
        \begin{tabular}{cccccc}
        \toprule
        Models   & Size &  1\% & 5\% & 10\% & 100\% \\
        \midrule
        IC-DST (Codex-davinci)\citep{hu-etal-2022-context}  & 175B    &  48.4 & 55.4 & 56.9 & 62.4 \\
        \sys-T5\baseM (ours)    & 220M    & 41.1  & 56.4 & 62.0 & 75.5 \\
        \bottomrule
        \end{tabular}
\end{table}

\paragraph{Comparison between in-context learning and finetuned models in the low-resource setting.}
\autoref{tab:icl_comparison} shows the comparison between in-context learning and full finetuned models. We use the same 1\%, 5\% and 10\% training set provided in \citet{hu-etal-2022-context}. \sys-T5\baseM surpasses Codex-davinci when the 5\% training ($\approx$ 374 examples) is available with 0.1\% model size. 
The result suggests that the small, finetuned model is still useful when a reasonable amount of training data is available.

\begin{table}[ht]
\caption{Comparison of training costs across models and the performance on \mwoz test set.} 
\centering
\small
    \begin{tabular}{cccc}
    \toprule
                  & Data                    & Training \flops $\downarrow$  & JGA on the test set$\uparrow$ \\\midrule
    T5\baseM      & $(\Xcal, \Ycal)$        &  $1.5 \times 10^{17}$         & 71.5           \\
    \pie-T5\baseM & $(\Xcal, Y_u, Z_u)$     &  $5.0 \times 10^{18}$         & 76.3           \\
    \pid-T5\baseM & $(\Xcal, \Ycal, \Zcal)$ &  $1.2 \times 10^{17}$         & 75.5                 \\\bottomrule
    \end{tabular}
\label{tab:training_efficiency}
\end{table}
\paragraph{Training efficiency.}
\autoref{tab:training_efficiency} shows the training costs associated with models and training strategies described in \autoref{subsec:training_procedure}. 
Vanilla T5 may have longer sequences in a batch, whereas \sys-T5 breaks the sequence into smaller subtasks, reducing padding issues.
\pie-T5\baseM incorporates prompts within its encoder, necessitating the enumeration of all prompts $\Zcal = (Z_1, \dots, Z_U)$ for every input $\Xcal$ during both training and testing phases. 
Consequently, \pie-T5\baseM requires more \flops, i.e., longer training duration, as the total number of training samples is increased by a factor of $U$. 
Conversely, \sys-T5\baseM allows the reuse of the same input across all prompts, keeping the total number of training examples the same as T5.
\sys-T5\baseM not only maintains a comparable JGA score and has efficient inference but also  reduces  training costs.

\begin{table}[htp!]
\caption{Comparison of latency (measured in msec) between different levels of GPUs on \mwoz test set.
L$_{\text{A100}}$ and L$_{\text{2080Ti}}$ stand for latency on NVIDIA A100 and RTX 2080Ti, respectively.
Sp represent the relative speed-up relative to \pie-T5\baseM or \pid-T5\largeM. 
} 
\centering
\small
    \begin{tabular}{cccccc}
    \toprule
    \multicolumn{1}{l}{} & JGA$\uparrow$  & L$_{\text{A100}}$ $\downarrow$ & Sp$_{\text{A100}}$ $\uparrow$  & L$_{\text{2080Ti}}$ $\downarrow$ & Sp$_{\text{2080Ti}}$ $\uparrow$      \\\midrule
    \pie-T5\baseM        & 76.3 & 146               & 1.0x     & 209                 & 1.0x     \\
    \pid-T5\baseM        & 75.5 & 78                & \bf1.9x  & 91                  & \bf2.3x     \\
    \midrule                           
    \pie-T5\largeM       & 77.5 & 413               & 1.0x     & 625                 & 1.0x     \\
    \pid-T5\largeM       & 76.5 & 147               & \bf2.8x  & 163                 & \bf3.8x     \\\bottomrule
    \end{tabular}
\label{tab:latency}
\end{table}
\paragraph{Latency on different levels of GPUs.}
\autoref{tab:latency} illustrates that our \sys-T5\baseM  surpasses \pie-T5\baseM in efficiency under a single-instance scenario. 
This difference becomes more pronounced when using a consumer-grade GPU, e.g., NVIDIA RTX 2080Ti, and inferencing on the larger model, showing \sys-T5 is more suitable when only lower resources are available.

\begin{table}[!htp]
    \caption{
    Comparison between different subtask scales, i.e., 30 domain slots or 5 domains, on \mwoz test set. 
    Sp represents the relative speed-up in latency computed on NVIDIA A100.
    The model with the domain subtask predicts all active slot values in that domain.}
    \centering
    \small
        \begin{tabular}{ccccc}
        \toprule
        Model        
        & Subtask
        & JGA $\uparrow$
        & \flops $\downarrow$ 
        & Sp $\uparrow$ 
        \\
        \midrule
        T5\baseM      & All         & 71.5 & 1.0x & 1.0x         \\
        \sys-T5\baseM & Domain      & 72.5 & 2.3x & 12.3x \\
        \sys-T5\baseM & Domain-Slot & 75.5 & 2.5x & 5.9x         \\
        \bottomrule
        \end{tabular}
    \label{tab:granularity}
\end{table}
\paragraph{Effect of different subtask granularity.}
In \mwoz, T5's output can be broken down into subtasks according to either domains (where the output is a sequence of observed slots and their values) or domain-slot pairs (where the output is the slot value). 
Each domain or domain-slot pair is associated with a individual prompt. As demonstrated in \autoref{tab:granularity}, \sys-T5\baseM reveals that employing multi-prompt decoding can enhance both the inference speed and the JGA score, regardless of the granularity of the subtask units. While utilizing domains as subtask units leads to more speed-up, the use of slots as subtask units yields the best JGA.

\begin{table}[!htp]
    \caption{Comparison between T5, \pie-T5 and \sys-T5 on \acibench test set.}
    \centering
    \small
        \begin{tabular}{cccccc}
        \toprule
                                &      & \multicolumn{4}{c}{Section (ROUGE-L)} \\\cmidrule(lr){3-6}
        Model                  & All  & 1 & 2 & 3 & 4 \\
        \midrule
        T5\baseM                & 47.9 & 34.3 & 28.8 & 28.4 & 17.9 \\
        \pie-T5\baseM           & 53.8 & \bf36.9 & 57.2 & 50.9 & 35.4 \\
        \sys-T5\baseM           & \bf54.0 & 36.6 & \bf57.7 & \bf58.9 & \bf35.9 \\
        \bottomrule
        \end{tabular}
    \label{tab:subtask_is_better}
\end{table}
\paragraph{Effect of subtasking for long output.}
In \acibench, We adopt the structure proposed by \cite{yim2023aci} for organizing the summary output into four distinct parts: subjective, objective examination, objective findings, and assessment with planning. The mean section output lengths are specified as 285, 98, 35, 254 tokens, respectively. The first section details the patient's medical history, while the fourth section focuses on assessment and planning; these sections surpass the second and third sections in terms of length.
\autoref{tab:subtask_is_better} reveals that the standard T5\baseM model underperforms with longer outputs, especially when the generation reaches the end of the sequence, i.e., performance of later sections are much worse than for the other models.
Both \pie-T5\baseM and \sys-T5\baseM models demonstrate improved performance with subtasking, allowing the model to ``focus'' on one subtask at a time.

\section{Related Work}
Improving the efficiency of transformer decoding hinges on minimizing memory access and reducing redundant computations via two primary ideas, increasing operational intensity and the reuse of key-value tensors.
While most current research focuses on decoder-only models, our work emphasizes encoder-decoder models, as they offer superior performance in specialized domains.

\paragraph{Attention optimizations.}
Increasing operational intensity can be achieved through hardware-friendly attention functions or model architectures, which increase the number of operational operations per memory access. 
FlashAttention \citep{dao2022flashattention} and FlashInfer \citep{flashinfer} use kernel-level optimization to fuse the attention mechanism into a single kernel function.
Multi-query attention \citep{Shazeer2019FastTD} and grouped-query attention \citep{ainslie-etal-2023-gqa} modify transformer architectures to employ a single (or fewer) key-value attention head for multiple query heads to reduce the memory access.

\paragraph{Key-value tensor caching.}
Another method to enhance decoding efficiency is the reuse of key-value tensors of shared common prefixes. 
This technique reduces redundant computations for subsequent requests with the same shared prefix.
Methods like paged attention \citep{kwon2023efficient} and radix attention \citep{zheng2023efficiently} mitigate the redundant storage of overlapping key-value cache.
While paged attention and radix attention address the memory fragmentation issue and enable memory reuse for shared prefix, 
they remain less than optimally efficient because their implementation lacks compute-level memory optimization. Consequently, the key-value cache of the shared prefixes still needs to be loaded multiple times during computation.

Although most aforementioned methods focus on decoder-only models, our method is compatible with these kernel-level efficiency techniques, e.g., FlashAttention \footnote{\texttt{FlashT5} enhances T5 by incorporating FlashAttention. \url{https://github.com/catie-aq/flashT5}} and paged attention \footnote{As of 05/22/2024, \href{https://github.com/vllm-project/vllm/pull/4837}{vLLM is planned to support encoder-decoder models but is still under development.}}, in principle leading to further efficiency gains when used in concert. 

Hydragen \citep{juravsky2024hydragen} is the most closely related concurrent research to our work. Both our method and Hydragen aim to increase operational intensity by sharing key-value tensors for shared prefixes. However, Hydragen mainly focuses on decoder-only models and emphasizes speed-up metrics in question answering. In contrast, we focus on smaller encoder-decoder models, which can outperform larger decoder-only models in specialized in-domain tasks. We evaluate both speedup and accuracy metrics across various applications, including dialogue state tracking, summarization, and question answering.
\section{Conclusion}
We study settings for decomposable tasks in NLP where multiple subtask prompts are applied to the same document or dialogue. 
The subtasking approach allows an encoder-decoder model to individually address smaller and simpler components of the main task, leading to improved task performance.
The strategy of moving the prompts from the encoder to the decoder allows our \sys configuration to reduce computational costs by encoding the input just once and then sharing it to multiple subtasks to decode outputs in parallel, which further speeds up inference time while either maintaining or improving task performance. 
Our approach achieves higher efficiency and comparable accuracy to existing approaches, which is particularly valuable in scenarios where computational resources are scarce.

\section*{Acknowledgments}
We extend our gratitude to Chen-Yu Ho for offering valuable suggestions and support in the implementation process. We also thank all members of the TIAL lab and NLP groups at the University of Washington who provided suggestions and insights into this work.  This work was supported in part by NSF IIS 2113530.
\bibliography{custom}

\bibliographystyle{plainnat}

\appendix
\section{Limitations}
\label{app:limitation}

The types of tasks where our method is applicable are currently limited because we require decomposable tasks with a shared input document.
The subtasking strategies in our datasets were designed by humans, e.g., according to structured output sections. Instead of using human-designed subtasking rules, a potential avenue for exploration is to allow a model to learn how to subtask, which can additionally make more tasks possible. While our subtasking experiments use only encoder-decoder models, our strategy of sharing an embedding and decomposing a task should work with decoder-only models, but experimental analysis is left to future work.

\section{Detailed Performance Analysis}
\label{app:comp-anal}

Higher operational intensity bring more efficient matrix computation in modern accelerators such GPUs/TPUs. In this section, we detail the performance analysis and compare the operational intensity ratios of \pie and \sys. To simplify equations, we follow the previous work \cite{Shazeer2019FastTD,de-jong-etal-2023-fido,ainslie-etal-2023-gqa}, using the inverse operational intensity ($\Rcal$) to compare all modules. The lower inverse ratio indicates higher operational intensity, hence better performance.

We discuss memory access and number of floating-point operations of encoder's self-attention, decoder's self-attention and decoder's cross-attention for \pie and \sys, respectively. 
We denote $n_s$, $n_t$ and $n_p$ as input length, output length and prompt length. $U$ is number of prompts/subtasks. $d$ is the joint dimension of all heads of key/query/value vectors.

We approximate the memory access and the number of operations based on the dominant terms in the self- and cross-attention and ignore the constant terms. 
For a single input, the memory access of the key or value tensors $\Mb$ or $\Nb$ are $n_s d$ and $n_t d$. 
The memory access of the projection matrices of key/query/value/output tensors $W^K$, $W^Q$, $W^V$, $W^O$ is $d^2$. 
As described in \citet{Shazeer2019FastTD}, the number of operations in both attention mechanisms is dominated by the matrix projections which are used to obtain projected query/key/value/output tensors ($\Qb$/$\Kb$/$\Vb$/$\Ob$); thus, the number of operations is approximated as $n_s d^2$ or $n_t d^2$.

\subsection{Encoder's self-attention}
The inverse operational intensity of \pie's encoder can be written as follows,
\begin{align*}
\Rcal^{\text{Enc-self}}_{\pie} 
= & \underbrace{Ub (n_s + n_p) d + d^2}_{\text{memory access}} \bigg/ \underbrace{U b (n_s + n_p) d^2}_{\text{\# operations}}\\
= & \frac{1}{d} + \frac{1}{Ub (n_s + n_p)},
\end{align*}
where \pie's encoder individually encodes $U$ prompts $(Z_1, \dots, Z_U)$ with input $\Xcal$. $\Rcal^{\text{Enc-self}}_{\pie}$ is a low ratio given the fact that $n_s$ is usually an hundred or a thousand tokens and $d$ is usually near a thousand.

Different from \pie's encoder, \sys's encoder only encodes input $\Xcal$ and leaves prompts in the decoder. Thus, the memory access is lower than \pie by a factor of $U$ and only the input length $n_s$ is considered. More formally the inverse operational intensity of \sys's encoder is denoted as
\begin{align*}
\Rcal^{\text{Enc-self}}_{\sys} 
= & \underbrace{b n_s d + d^2}_{\text{memory access}} \bigg/ \underbrace{b n_s d^2}_{\text{\# operations}}\\
= & \frac{1}{d} + \frac{1}{b n_s}.
\end{align*}
Again, $n_s$ and $d$ is around a thousand, resulting in $\Rcal^{\text{Enc-self}}_{\sys}$ is also a low ratio. Compared to \pie, \sys requires fewer number of operations, saving more memory usage and enabling faster computation.

\subsection{Decoder's self-attention}

In \pie, prompts are encoded in the encoder, the decoder only accounts for generating target tokens. The inverse operational intensity of \pie encoder's self-attention is denoted as
\begin{align*}
\Rcal^{\text{Dec-self}}_{\pie} 
= & \underbrace{Ub n^2_t d + n_t d^2}_{\text{memory access}} \bigg/ \underbrace{U b n_t d^2}_{\text{\# operations}}\\
= & \underbrace{\frac{n_t}{d}}_{\text{dominant term}} + \frac{1}{Ub},
\end{align*}
where $\frac{n_t}{d}$ is the dominant term that causes the issue of slower incremental decoding.

\sys's decoder encodes prompts and incrementally generates output tokens. We decouple the analysis of encoding prompts and generating output tokens since the prompts are all given whereas the output tokens are incrementally decoded. The decoder just need to encode the prompts once; as a result, the ratio can be written as
\begin{align*}
\Rcal^{\text{Dec-self,prompt}}_{\sys} 
&= \underbrace{U b n_p d + d^2}_{\text{memory access}} \bigg/ \underbrace{U b n_p d^2}_{\text{\# operations}}\\
&= \frac{1}{d} + \frac{1}{Ubn_p}
\end{align*}
Obviously, the $\Rcal^{\text{Dec-self,prompt}}_{\sys}$ is a low ratio; thus the encoding prompts in the decoder part is efficient.
In addition to encoding prompts, \sys's decoder needs to generate output tokens. The ratio $\Rcal^{\text{Dec-self,output}}_{\sys}$ is the same as $\Rcal^{\text{Dec-self}}_{\pie}$.

\subsection{Decoder's cross-attention}

In transformer architecture, the decoder's cross-attention is the key issue that cause the computation inefficiency since the inference is incremental and cross-attention needs to read the huge chunk of key and value cached tensors from the encoder. The inverse operational intensity of \pie decoder's cross-attention can be denoted as follows,
\begin{align*}
\Rcal^{\text{Dec-cross}}_{\pie}
= & \frac{\overbrace{Ub (n_s + n_p) n_t d + U b n_t d + n_t d^2}^{\text{memory access}}}{\underbrace{U b n_t d^2}_{\text{\# operations}}}\\
= & \underbrace{\frac{n_s+n_p+1}{d}}_{\text{dominant term}} + \frac{1}{Ub},
\end{align*}
where the prompts are encoded with the input in the encoder; hence the length of cached tensors is $n_s + n_p$. The dominant term results in a serious bottleneck especially when long input $n_s$ is fed into the model.

Similar to the \sys decoder's self-attention, we consider the cross-attention on prompts and output separately. The ratio for encoding prompts is as follows,
\begin{align*}
\Rcal^{\text{Dec-cross,prompt}}_{\sys} 
= & \frac{\overbrace{b n_s d + Ubn_pd + d^2}^{\text{memory access}}}{\underbrace{Ub n_p d^2}_{\text{\# operations}}}\\
= & \underbrace{\frac{1}{d} \cdot \left(\frac{n_s}{Un_p}+1\right) }_{\text{dominant term}} + \frac{1}{Ubn_p}.
\end{align*}
In the dominant term, the input length $n_s$ is divided by the factor of $U n_p$.

On the other hand, the ratio of decoding output tokens is
\begin{align*}
\Rcal^{\text{Dec-cross,output}}_{\sys} 
= & \frac{\overbrace{b n_s n_t d + U b n_t d + n_t d^2}^{\text{memory access}}}{\underbrace{U b n_t d^2}_{\text{\# operations}}}\\
= & \underbrace{\frac{1}{d} \cdot \left( \frac{n_s}{U} + 1 \right)}_{\text{dominant term}} + \frac{1}{Ub}.
\end{align*}
Similarly, in the dominant term, the input length $n_s$ is divided by the factor of $U$. Overall, in \sys decoder's cross-attention, the dominant term of the incremental decoding is reduced by a factor of $U$ or $Un_p$ since \sys shares the same input key and value cached tensors and broadcast the matrix operations while performing the cross-attention.

\begin{table*}[h]
\caption{The \sys-T5 hyperparameters used in the training and testing.}
\centering
\small
\begin{tabular}{ccccccc}
\toprule
                                  & \multicolumn{2}{c}{\mwoz}              & \multicolumn{2}{c}{\acibench}     & \multicolumn{2}{c}{\radqa}            \\
                                  & \sys-T5\baseM           & \sys-T5\largeM         & \sys-T5\baseM         & \sys-T5\largeM      & \sys-T5\baseM         & \sys-T5\largeM          \\
                                \midrule
Batch size                        & 4                 & 1                  & 1                 & 2             & 4                & 2                 \\
Grdient accumulation              & 64                & 32                 & 16                & 16            & 16               &32 \\
Effective batch size              & 64                & 64                 & 32                & 32            & 64               & 64                \\
\# epochs                         & 6                 & 4                  & 100               & 100           & 50               & 15                \\
Max input length                  & 1024              & 1024               & 3072              & 3072          & 1024             & 1024              \\
Max output length                 & 24                & 24                 & 1024              & 1024          & 92               & 92                \\
Max prompt length                 & 8                 & 8                  & 6                 & 6             & 36               & 36                \\
\# outputs                        & 30                & 30                 & 4                 & 4             & 2-6              & 2-6               \\
\# beams                          & 1                 & 1                  & 4                 & 4             & 1                & 1              \\\bottomrule
\end{tabular}
\label{tab:training_details}
\end{table*}

\section{Training Details}
\label{app:training_details}
\autoref{tab:training_details} presents the hyperparameters selected for the training and testing phases. We executed a search for the optimal learning rate across the following set of values: $\{5 \times 10^{-4}, 3 \times 10^{-4}, 1 \times 10^{-4}, 7 \times 10^{-5}, 5\times 10^{-5}, 3 \times 10^{-5}\}$ to identify the most effective learning rate for each dataset. 
All reported values represent the medians of three different random runs.
The remaining hyperparameters not mentioned in \autoref{tab:training_details} are set to the default values provided by the HuggingFace Transformers package.
Training time varies because of dataset size, model size, model configuration and training procedure. 
Our experiments, which utilize T5\baseM as the primary framework, are carried out using a single NVIDIA A40.
Training T5\baseM and \sys-T5\baseM on the \mwoz dataset typically requires approximately 5 GPU hours, whereas it takes around 46 GPU hours for \pie-T5\baseM. In the case of \acibench, where the dataset is relatively small, T5\baseM, \pie-T5\baseM, and \sys-T5\baseM require roughly 4 GPU hours each. On the other hand, for \radqa, \pie-T5\baseM takes 2 hours, while \pid-T5\baseM requires 3 GPU hours.
When switching to T5\largeM, the required GPU training time increases by a factor of 2 to 3 times compared to the T5\baseM models.

We use t5-base\footnote{\url{https://huggingface.co/google-t5/t5-base}} and t5-large\footnote{\url{https://huggingface.co/google-t5/t5-large}} checkpoints downloaded from HuggingFace as initialization for \mwoz and \acibench. For \radqa, we follow the previous work \cite{pmlr-v209-eric23a} to use pretrained clinical T5 models.\footnote{\url{https://physionet.org/content/clinical-t5/1.0.0/}}

\begin{table*}[ht]
\centering
\small
\caption{Statistics are calculated on the each full data set. Input and output lengths are calculated based on Huggingface T5 tokenizer.}
\begin{tabular}{lccc}
\toprule
                    & \mwoz                 & \acibench               & \radqa                \\\midrule
Data                & Task oriented         & Medical                 & Medical               \\
Input type          & Dialogue              & Dialogue                & Document              \\
Task                & Dialog state tracking & Summarization           & Question Answering    \\
\# examples         & 9887                  & 207                     & 6148                  \\
Input length        & 289$_{\pm\text{108}}$ & 1725$_{\pm\text{511}}$  & 137$_{\pm\text{157}}$ \\
Output length       & 56$_{\pm\text{26}}$   & 693$_{\pm\text{200}}$   & 28$_{\pm\text{49}}$   \\
Prompt type         & Fixed                 & Fixed                   & Free-form              \\
\# prompts          & 30 slots or 5 domains & 4 sections              & --                     \\\bottomrule
\end{tabular}
\label{tab:datasets_comparison}
\end{table*}
\section{Datasets}
\label{app:datasets}
In terms of data preprocessing, we follow the previous works \cite{ye-etal-2022-assist, yim2023aci, pmlr-v209-eric23a} to process \mwoz, \acibench and \radqa, respectively. The dataset statistics are shown in \autoref{tab:datasets_comparison}.

\section{License of Artifacts}
\label{app:license}
The licensing for the code from HuggingFace's transformers \cite{wolf-etal-2020-transformers} falls under the Apache License, Version 2.0. 
PyTorch \cite{paszke2019pytorch} is open-source software released under the modified BSD license.
Meanwhile, the calflops \cite{calflops} code is protected under the MIT License. 
Detailed conditions for utilizing our artifacts will be provided within the package we distribute.
\end{document}